\documentclass[letterpaper]{article} 
\usepackage{aaai2026}  
\usepackage{times}  
\usepackage{helvet}  
\usepackage{courier}  
\usepackage[hyphens]{url}  
\usepackage{graphicx} 
\usepackage{float}
\usepackage{natbib}  
\usepackage{caption} 
\usepackage{times}  
\usepackage{helvet}  
\usepackage{courier}  
\usepackage[hyphens]{url}  
\usepackage{graphicx} 
\usepackage{booktabs}
\usepackage{cite}
\usepackage{amsmath,amssymb,amsfonts}
\usepackage{algorithmic}
\usepackage{graphicx}
\usepackage{booktabs}
\usepackage{xcolor} 
\usepackage{colortbl} 
\usepackage{textcomp}
\usepackage{algorithm}
\usepackage{algorithmic}

\usepackage{newfloat}
\usepackage{listings}
\DeclareCaptionStyle{ruled}{labelfont=normalfont,labelsep=colon,strut=off} 
\floatstyle{ruled}
\newfloat{listing}{tb}{lst}{}
\floatname{listing}{Listing}
\title{Towards Authentic Movie Dubbing with \\ Retrieve-Augmented Director-Actor Interaction Learning}

\author {
    Rui Liu\thanks{Corresponding author.}\equalcontrib,
    Yuan Zhao\equalcontrib,
    Zhenqi Jia\\
}
\affiliations {
    College of Computer Science, Inner Mongolia University, Hohhot, China\\
    imucslr@imu.edu.cn, 
    zy404nf@163.com, 
    jiazhenqi7@163.com
}

\usepackage{bibentry}

\begin{document}

\maketitle

\begin{abstract}
The automatic movie dubbing model generates vivid speech from given scripts, replicating a speaker's timbre from a brief timbre prompt while ensuring lip-sync with the silent video.
Existing approaches simulate a simplified workflow where actors dub directly without preparation, overlooking the critical director–actor interaction. In contrast, authentic workflows involve a dynamic collaboration: directors actively engage with actors, guiding them to internalize the context cues, specifically emotion, before performance.
To address this issue, we propose a new Retrieve-Augmented Director-Actor Interaction Learning scheme to achieve authentic movie dubbing, termed \textbf{Authentic-Dubber}, which contains three novel mechanisms: 
(1) We construct a multimodal Reference Footage library to simulate the learning footage provided by directors. Note that we integrate Large Language Models (LLMs) to achieve deep comprehension of emotional representations across multimodal signals.
(2) To emulate how actors efficiently and comprehensively internalize director-provided footage during dubbing, we propose an Emotion-Similarity-based Retrieval-Augmentation strategy. This strategy retrieves the most relevant multimodal information that aligns with the target silent video.
(3) We develop a Progressive Graph-based speech generation approach that incrementally incorporates the retrieved multimodal emotional knowledge, thereby simulating the actor's final dubbing process.
The above mechanisms enable the Authentic-Dubber to faithfully replicate the authentic dubbing workflow, achieving comprehensive improvements in emotional expressiveness. Both subjective and objective evaluations on the V2C-Animation benchmark dataset validate the effectiveness. The code and demos are available at \textcolor[rgb]{0.93,0.0,0.47}{\url{https://github.com/AI-S2-Lab/Authentic-Dubber}}.
\end{abstract}

\begin{figure}[h]
    \centering
    \includegraphics[width=\linewidth]{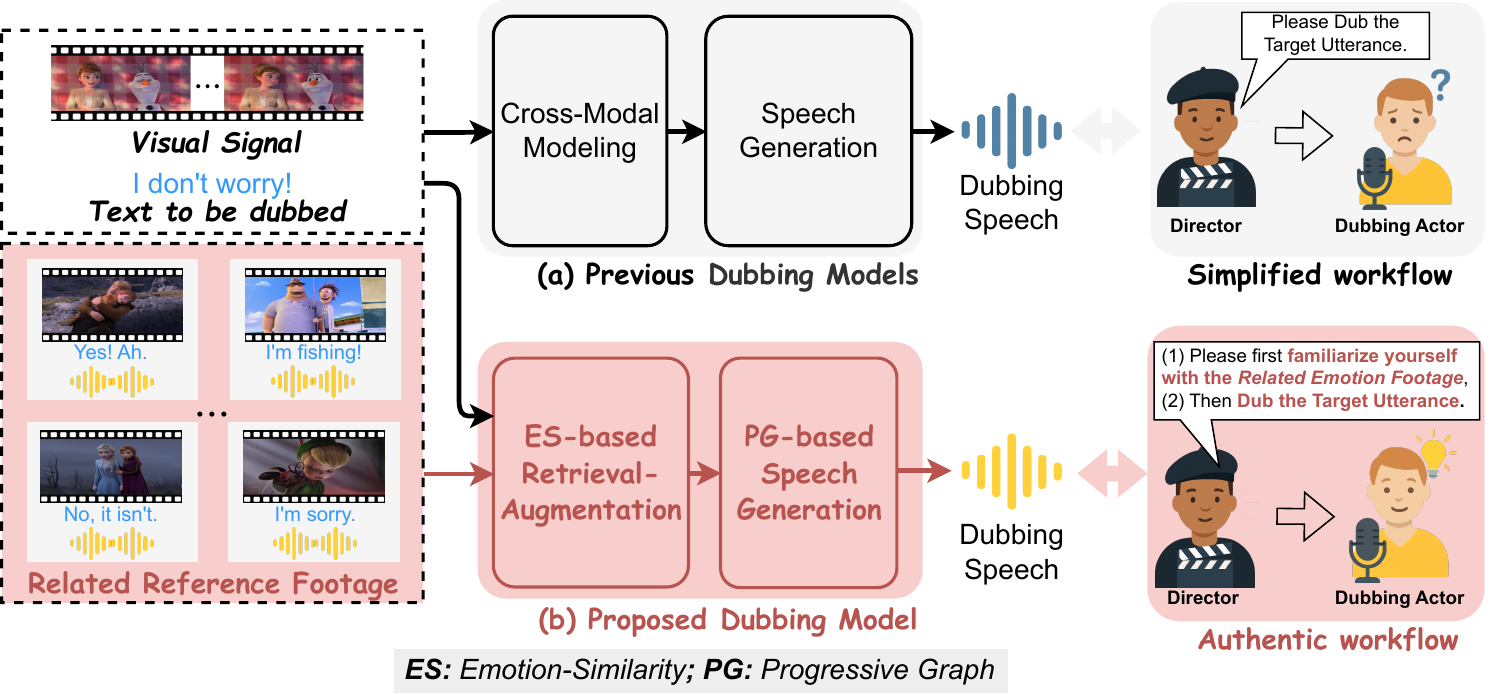}

    \caption{(a) Previous models rely solely on cross-modal modeling of the target utterance to generate speech, which results in limited emotional expressiveness. (b) Our method enables expressive dubbing through three mechanisms: Multimodal Reference Footage Construction, Emotion-Similarity-based Retrieval-Augmentation, and Progressive Graph-based Speech Generation.}
    \vspace{-4mm}
    \label{fig_1}
\end{figure}

\section{Introduction}

Movie Dubbing, also known as Visual Voice Cloning (V2C) \cite{V2CNET, speaker2dubber, li2025dubbing}, aims to generate vivid speech from given scripts, replicating a speaker's timbre from a brief timbre prompt while ensuring lip-sync with the silent video. V2C holds significant potential and value for applications in movie production and commercial Artificial Intelligence Generated Content \cite{AIGC}.

Traditional dubbing work mainly focuses on improving pronunciation quality \cite{cong2024styledubber, speaker2dubber, cong2025emodubber}, audio-visual synchronization \cite{hu2021neural, cong2025emodubber, voicecraft_dubber, alidit, lu2022visualtts}, and expressiveness \cite{HPMDubbing, zhao2025m2ci_dubber, li2025fccondubber, zhao2024mcdubber, MM-MovieDubber}. To improve pronunciation quality, Speaker2Dubber \cite{speaker2dubber} designs a multi-task pre-training to learn pronunciation knowledge before using the dubbing dataset. To enhance audio-visual synchronization, FlowDubber \cite{flowdubber} introduces a dual contrastive learning approach between lip movement sequences and phoneme sequences. For improving expressiveness, ProDubber \cite{zhang2025produbber} proposes a two-stage approach. It includes prosody-enhanced acoustic pre-training and acoustic-disentangled prosody adaptation, ensuring high audio quality and accurate prosody alignment. The above works pave the way for accelerated advancement in movie dubbing technologies.

Despite the progress, these methods simulate a simplified workflow and overlook the critical interaction between the director and the actor, which limits their ability to fully model dubbing expressiveness, especially emotional expression. Specifically, previous dubbing models exhibit limited emotional expression due to their reliance solely on cross-modal modeling of the target utterance, as shown at the top of Fig. \ref{fig_1}. These models simulate a simplified workflow in which actors proceed directly to perform dubbing, facing confusion without preparation. They overlook the crucial interaction between the director and the actors. In authentic movie dubbing, directors typically require dubbing actors to first become sufficiently familiar with emotional reference footage, which contains rich knowledge of emotional expression. This process helps actors internalize contextual cues—particularly emotional ones—before the actual performance, as illustrated at the bottom of Fig. \ref{fig_1}. Only after thoroughly studying the footage and cumulatively acquiring this emotional knowledge can the dubbing actor perform emotionally expressive dubbing. Through this interactive learning process between the director and the actor, the final generated dubbing can exhibit rich emotional expression.

To address the above issue, we propose a novel Retrieval-Augmented Director-Actor Interaction Learning scheme for authentic movie dubbing, termed \textbf{Authentic-Dubber}, which consists of three novel mechanisms: 
(1) To simulate the learning footage provided by directors, we construct a Multimodal Reference Footage Library (MRFL). We design specialized emotion extractors for indirect multimodal emotional information, including the scene's emotional atmosphere, the character's facial expression changes, and the script's emotional semantics, as well as their matched direct emotional audio within each movie clip. In this process, we integrate Large Language Models (LLMs) to extract emotional captions to enable deep comprehension of emotional representations across multimodal signals. 
(2) To emulate how actors efficiently and comprehensively internalize director-provided footage during dubbing, we propose an Emotion-Similarity-based Retrieval-Augmentation (ESRG) strategy. ESRG uses the target utterance's basic emotion, i.e., scene, face, and text, as separate queries. Each query retrieves similar indirect emotion information and matched direct emotional audio from the MRFL. These retrieved items serve as the most relevant multimodal information aligned with the target dubbing video. (3) To simulate the actor's final dubbing process, we propose a Progressive Graph-based Speech Generation (PGSG) approach. This method incrementally learns emotional knowledge from the target's basic emotional source, the retrieved indirect multimodal information, and the direct emotional audio. PGSG follows a progressive construct-and-encode paradigm over the Basic Emotion Graph, Indirect Emotion Extended Graph, and Direct Emotion Extended Graph. Finally, the emotional knowledge acquired from these three stages is aggregated in an Emotion Knowledge-based Speech Synthesizer to generate emotionally expressive speech. The main contributions of this paper are as follows:
\begin{itemize}
\vspace{-1mm}
    \item Based on the dubbing workflow in real-world scenarios, we propose, for the first time, a movie dubbing model, termed \textit{Authentic-Dubber}, that simulates authentic workflows, aiming to enhance the emotional expressiveness of movie dubbing.
    \vspace{-1mm}
 \item We adopt three key technologies, that are \textit{Multimodal Reference Footage Construction}, \textit{Emotion-Similarity-based Retrieval Augmentation}, and \textit{Progressive Graph-based Speech Generation}, to simulate the interaction process between directors and actors.
 \vspace{-1mm}
 \item Subjective and objective experiments on benchmark datasets, along with additional detailed analytical experiments, demonstrate the effectiveness of our method in improving emotional expressiveness.
\end{itemize}

\section{Related Works}

\begin{figure*}[ht]
    \centering
    \includegraphics[width=0.8\linewidth]{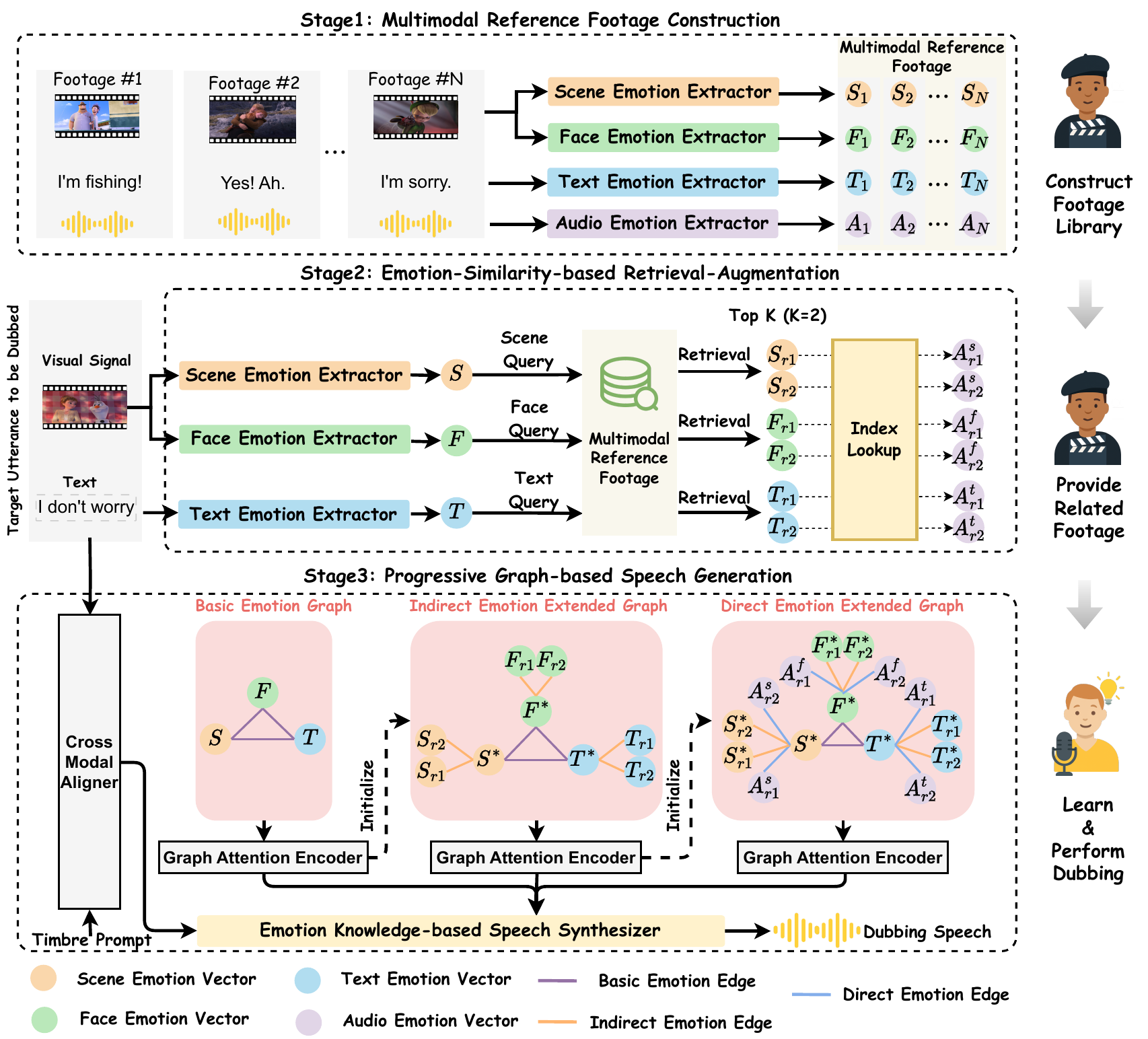}
    \caption{The proposed Authentic-Dubber consists of Multimodal Reference Footage Construction, Emotion-Similarity-based Retrieval-Augmentation, and Progressive Graph-based Speech Generation. (* means  that the node's initial vector representation is initialized from the immediately preceding graph.)}
    \vspace{-3mm}
    \label{fig_2}
\end{figure*}

\subsection{Retrieval-Augmented Generation }
Retrieval-Augmented Generation (RAG) compensates for the limitations of a single model’s knowledge by leveraging rich and explainable additional knowledge \cite{kang2024retrieval, yuan2024retrieval, yang2024rag, ghosh2024recap} and has demonstrated strong performance in related multi-modal information processing tasks. For example, $\text{CM}^2$ \cite{kim2024you} improves image caption generation by using cross-modal video-text matching to retrieve prior knowledge from an external memory bank. Re-Imagen \cite{chen2022RE_imagen} retrieves relevant (image, text) pairs from an external multi-modal knowledge bank, allowing high-fidelity image generation. 
Note that RAG methods provide an excellent example of addressing knowledge-intensive tasks \cite{gao2023retrieval, lewis2020retrieval, guu2020retrieval}. Inspired by the above ideas, we believe that in the dubbing process, the reference footage provided by directors serves as an invaluable source of knowledge for actors. Therefore, we propose an emotion-similarity-based retrieval-augmentation strategy to simulate how actors learn from such materials. However, unlike previous RAG schemes that retrieve knowledge through embedding-based similarity calculation and directly feed the retrieved features into downstream modules, our Authentic-Dubber incorporates the following special designs tailored to movie dubbing: 1) We calculate the similarity of diverse emotional expressions—such as scenes, facial expressions, texts, and speech—from the constructed learning materials to identify relevant samples; 2) We develop a hierarchical knowledge fusion strategy based on a progressive graph, which integrates both the utterance to be dubbed and the retrieved multimodal signals. This has not been considered at all in previous dubbing efforts.

\subsection{Graph-Based Knowledge Modeling}
Graph Neural Network (GNN) is designed to process graph-structured data \cite{sanchez2021gentle}, capturing relationships and structural information by learning representations of nodes and edges \cite{hamilton2017representation, hamilton2020graph}.
In movie dubbing and related tasks, several works employ GNN to model complex interactions between multimodal signals within utterances \cite{liu2024emotion}. For example, M2CI-Dubber \cite{zhao2025m2ci_dubber} uses a graph attention network to model the relation between the multiscale multimodal context and the target utterance, enhancing prosody expressiveness. Li et al. \cite{lijingbei_dialoguegnn_gcn} use GNN to model both intra-speaker and inter-speaker dependencies, enhancing the speaking style of the generated speech. 
While existing approaches also leverage GNN to encode emotional knowledge within utterances, our method innovatively introduces a progressive graph fusion mechanism. The key distinction from prior work lies in (1) We propose a novel progressive construct-and-encode paradigm: first encoding a graph of the target basic emotion, then incorporating retrieved multimodal emotional information for further and indirect emotional modeling, and finally integrating emotional audio for direct emotion modeling. This gradual accumulation of emotional knowledge enhances the model’s emotion understanding.
(2) We introduce basic emotion edge, indirect emotion edge, and direct emotion edge to capture relationships among emotional nodes, boosting the model’s emotion modeling capabilities in dubbing.

\section{Authentic-Dubber: Methodology}
\subsection{Overview}
Given a script, a silent video clip, and a timbre prompt, the goal of \textbf{Authentic-Dubber} is to generate a speech with emotion expressiveness. The main architecture of the proposed model is shown in Fig. \ref{fig_2}. (1) We first construct a Multimodal Reference Footage Library (MRFL) based on the V2C \cite{V2CNET} dataset. For each sample, modality-specific emotion extractors are designed for indirect multimodal emotion information (scene, face, text), and matched direct emotion audio to extract their respective emotion vectors.  (2) In Emotion-Similarity-based Retrieval-Augmentation, the target utterance's basic emotion is individually encoded, and then the basic emotion is used separately as queries to retrieve indirect multimodal emotion information and direct emotional audio from MRFL. (3) In Progressive Graph-based Speech Generation, we adopt a progressive construct-and-encode paradigm over the \textit{Basic Emotion Graph, Indirect Emotion Extended Graph, and Direct Emotion Extended Graph} to accumulate emotional knowledge. Finally, an Emotion Knowledge-based Speech Synthesizer integrates the learned emotional knowledge to generate emotionally expressive dubbing while using the cross-modal alignment results of the text, timbre prompt, and visual frames as input.

\subsection{Multimodal Reference Footage Construction}

To simulate the learning footage provided by
directors, we first build a Multimodal Reference Footage Library (MRFL) based on the V2C dataset \cite{V2CNET}. For each sample \(i\), we design four emotion extractors for both the indirect emotional multimodal information and the direct emotional audio, projecting them into their respective emotional spaces.

\textbf{Scene Emotion Extractor.} Inspired by the general understanding capabilities of Large Language Models (LLMs), we first employ a video understanding model, VideoLLaMA 2 \cite{cheng2024videollama2}, to generate a scene emotion caption, while incorporating global low-level visual features such as hue, lightness, and saturation that influence emotion perception into the instruction prompts. The generated caption is then passed to a RoBERTa-based Text Emotion Recognition model (RTER) \footnote{https://huggingface.co/j-hartmann/emotion-english-roberta-large} \cite{EERL} to extract the scene emotion vector \( S_i \).

\textbf{Face Emotion Extractor.} Using VideoLLaMA 2 \cite{cheng2024videollama2}, we generate a caption describing facial emotion changes in the video, and the caption is then processed by RTER to produce the face emotion vector \( F_i \).

\textbf{Text Emotion Extractor.} We first use RTER to obtain a text-self emotion vector \( T_i^{self} \) from the input text. Considering that prior commonsense reactions~\cite{deng2023cmcu} influence the understanding of the text's emotion, we use COMET\footnote{\url{https://huggingface.co/svjack/comet-atomic-en}} to generate a reaction caption, which is also processed by RTER to generate the text-react emotion vector \( T_i^{react} \). Finally, the text-self emotion vector \( T_i^{self} \) is concatenated with the text-react emotion vector \( T_i^{react} \) to form the final text emotion vector \( T_i \).

\textbf{Audio Emotion Extractor.} We employ a universal emotion representation model, Emotion2Vec   \cite{ma2023emotion2vec} to extract the audio emotion vector \( A_i \).

\subsection{Emotion-Similarity-based Retrieval-Augmentation}
In this work, we focus on animated movie dubbing, where characters are virtually created and speaker-specific reference footage is often limited. To address this challenge, we introduce a speaker-agnostic retrieval strategy, which allows access to a broader and more emotionally diverse set of footage. Specifically, we design a separate retrieval mechanism based on different emotional knowledge. Considering that the target speech is absent in real-world movie dubbing scenarios, we utilize the target utterance's basic emotion to retrieve indirect multimodal emotion information, which in turn helps match the corresponding direct emotional audio via index lookup. For scene retrieval, the Scene Emotion Extractor generates the target utterance's scene emotion vector \(S\), which is compared via cosine similarity with the scene vectors in MRFL to retrieve the Top-\(K\) indirect scene emotion information \(S_{r1 \rightarrow rk}\) and the matched direct emotional audio \(A_{r1 \rightarrow rk}^{s}\). For face retrieval, the Face Emotion Extractor generates the target utterance's face emotion vector \(F\). Similar to scene retrieval, the Top-\(K\) indirect facial emotion information \(F_{r1 \rightarrow rk}\) and matched direct emotional audio \(A_{r1 \rightarrow rk}^{f}\) are retrieved. Notably, for text retrieval, the Text Emotion Extractor generates the target utterance’s text emotion vector \(T\) by concatenating the text-self emotion vector \(T^{self}\) and the text-react emotion vector \(T^{react}\). We compute the cosine similarity between \(T^{self}\) and \(T_i^{self}\), as well as between \(T^{react}\) and \(T_i^{react}\) in MRFL, and use the average of these values as the retrieval criterion for text retrieval. The Top-\(K\) indirect text emotion information \(T_{r1 \rightarrow rk}\) and matched direct emotional audio \(A_{r1 \rightarrow rk}^{t}\) are retrieved.
\subsection{Progressive Graph-based Speech Generation}
Inspired by the authentic dubbing workflow, where actors first understand the basic emotion of the target silent video and then refer to similar movie clips containing indirect multimodal emotional information and their matched direct emotional audio. Therefore, we design a Progressive Graph-based Speech Generation module, which accumulates learn emotional knowledge. It adopts a progressive construct-and-encode paradigm over the Basic Emotion Graph, Indirect Emotion Extended Graph, and Direct Emotion Extended Graph.

\textbf{Basic Emotion Graph.} Since the emotional knowledge of the target utterance is most closely related to the emotional expression of speech, we first guide the model to focus on learning it. Specifically, we construct a Basic Emotion Graph \( \mathcal{G}_{beg} \), where the nodes consist of the scene emotion vector \(S\), face emotion vector \(F\), and text emotion vector \(T\). To better capture the relationships between different basic emotion sources, we connect the emotion sources \( S \), \( F \), and \( T \) in a pairwise manner. Next, we utilize a Graph Attention Encoder (GAE) to encode \( \mathcal{G}_{beg} \), where the resulting graph \( \tilde{\mathcal{G}_{beg}} \) represents the learned emotional knowledge of the target basic emotion source.

\textbf{Indirect Emotion Extended Graph.} Based on the encoded \( \tilde{\mathcal{G}_{\mathrm{beg}}} \) and the retrieved indirect multimodal emotion nodes, we construct and initialize an Indirect Emotion Extended Graph \( \mathcal{G}_{\mathrm{ieg}} \). The retrieved nodes are connected to the basic emotion source nodes of the same modality. The encoded \( \tilde{\mathcal{G}}_{\mathrm{ieg}} \) further accumulatively learns emotional knowledge derived from the retrieved indirect emotion information.

\textbf{Direct Emotion Extended Graph.} Based on the encoded \( \tilde{\mathcal{G}_{\mathrm{ieg}}} \) and the matched emotion audio nodes, we construct and initialize a Direct Emotion Extended Graph \( \mathcal{G}_{\mathrm{deg}} \). The matched direct emotion audio is added as new nodes and connected to the basic emotion source from which the corresponding query is issued. The encoded \( \tilde{\mathcal{G}}_{\mathrm{deg}} \) continuously learns the emotional knowledge derived from the retrieved matched direct emotion audio.

\textbf{Emotion Knowledge-based Speech Synthesizer.}
Our emotion knowledge-based speech synthesizer performs hierarchical aggregation of learned emotional knowledge to enhance the emotional expressiveness of the generated speech. Meanwhile, it takes as input the output $H_{t,v,r}$ from the Cross-Modal Aligner. This aligner follows the architecture of StyleDubber \cite{cong2024styledubber}, achieving audio-visual synchronization based on the input script and visual frames, and learns voice from the timbre prompt. Specifically, the nodes in graphs $\tilde{\mathcal{G}}_{beg}$, $\tilde{\mathcal{G}}_{ieg}$, and $\tilde{\mathcal{G}}_{deg}$ are denoted as $H_{beg}$, $H_{ieg}$, and $H_{deg}$, respectively. The emotion aggregation is then performed as follows:
\begin{equation} \label{eq:emotion_aggregation_compact}
\begin{aligned}
    E_{t,v,r}^{beg} &= \text{Conv1D}\left( \left[H_{t,v,r};\ \text{CA}(H_{t,v,r}, H_{beg}, H_{beg}) \right] \right) \\
    E_{t,v,r}^{ieg} &= \text{Conv1D}\left( \left[E_{t,v,r}^{beg};\ \text{CA}(E_{t,v,r}^{beg}, H_{ieg}, H_{ieg}) \right] \right) \\
    E_{t,v,r}^{deg} &= \text{Conv1D}\left( \left[E_{t,v,r}^{ieg};\ \text{CA}(E_{t,v,r}^{ieg}, H_{deg}, H_{deg}) \right] \right) \\
    E_{t,v,r}^{out} &= \text{Conv1D}\left( \left[H_{t,v,r};\ E_{t,v,r}^{deg} \right] \right)
\end{aligned}
\end{equation}
where CA denotes Cross-Attention. Finally, the aggregated emotional representation \( E_{t,v,r}^{\text{out}} \) is passed to the Mel decoder to produce a Mel spectrogram, which is then converted into the final speech with rich emotion expression using a Vocoder \cite{liu2025multi, he2025multi-icassp}\footnote{\url{https://huggingface.co/nvidia/bigvgan_v2_44khz_128band_256x}}.

\section{Experimental Setup}

\subsection{Dataset}
V2C-Animation \cite{V2CNET} is a multi-speaker dataset designed for animated movie dubbing. It is collected from 26 Disney animated movies and includes 153 characters. The dataset consists of 10,217 video clips with paired audio and scripts, and is split into 60\% for training, 10\% for validation, and 30\% for testing. Notably, V2C is currently the only publicly available movie dubbing dataset with emotion annotations. Therefore, we evaluate our model solely on the V2C dataset to assess its effectiveness in enhancing emotional expressiveness in dubbing.

\subsection{Implementation Details}
Video frames are sampled at 25 frames per second, and all audio samples are resampled to 22.05 kHz. In the Short-Time Fourier Transform, the window length, frame size, and hop length are set to 1024, 1024, and 256, respectively. In Progressive Graph-based Speech Generation, the dimensionality of all input emotional features is set to 256 through a linear layer. 
The output dimension of the Graph Attention Encoder is set to 256. In the Emotion Knowledge-based Speech Synthesizer, the output dimension of the Conv1D layer is set to 256. During training, we use the Adam optimizer with parameters set to $\beta_1 = 0.9$, $\beta_2 = 0.98$, and $\epsilon = 10^{-9}$, with a learning rate of 0.00625. Both training and inference are implemented using PyTorch on an A800 GPU.

\begin{table*}[ht]
\setlength{\abovecaptionskip}{3pt} 
\setlength{\belowcaptionskip}{-3pt}
\centering

\renewcommand{\arraystretch}{1.2} 
\setlength{\tabcolsep}{6pt}
\resizebox{\textwidth}{!}{%
\begin{tabular}{lcccccc}
\toprule
Methods & EMO-ACC ($\uparrow$) & WER ($\downarrow$) & SECS ($\uparrow$) & MCD-DTW-SL ($\downarrow$) & MOS-DE ($\uparrow$) & MOS-SE ($\uparrow$) \\
\midrule
Ground-Truth & 99.96 & 22.03 & 100.00 & 0.00 & 4.416 ± 0.035 & 4.497 ± 0.044 \\
\hline
FastSpeech2 \cite{fs2} (ICLR 2021) & 42.39 & 33.30 & 25.47 & 14.72 & 3.058 ± 0.077 & 3.063 ± 0.082 \\
V2C-Net \cite{V2CNET} (CVPR 2022) & 43.07 & 67.98 & 40.65 & 19.16 & 3.146 ± 0.062 & 3.149 ± 0.064 \\
HPMDubbing \cite{HPMDubbing} (CVPR 2023) & 43.94 & 135.72 & 34.11 & 12.64 & 3.362 ± 0.049 & 3.320 ± 0.040 \\
StyleDubber \cite{cong2024styledubber} (ACL 2024) & 45.73 & 24.70 & 83.46 & \textbf{9.40} & 3.676 ± 0.048 & 3.738 ± 0.049 \\
Speaker2Dubber \cite{speaker2dubber} (MM 2024) & 44.55 & \textbf{18.27} & 81.26 & 9.82 & 3.432 ± 0.069 & 3.461 ± 0.069 \\
\hline
\textbf{Authentic-Dubber}  & \textbf{47.21} & 25.95 & \textbf{84.40} & 9.68 & \textbf{3.792 ± 0.055} & \textbf{3.889 ± 0.053} \\
\bottomrule
\end{tabular}
}

\caption{Objective and subjective (with 95\% confidence interval) evaluation results with other methods. $\uparrow$ ($\downarrow$) indicates that a higher (lower) value is better, and \textbf{bold} indicates the best score. The Authentic-Dubber significantly outperforms the baselines on emotion expressiveness.}

\label{table1}
\end{table*}

\begin{table}[ht]
\setlength{\abovecaptionskip}{3pt} 
\setlength{\belowcaptionskip}{-3pt}

\centering
\renewcommand{\arraystretch}{1.2} 
\setlength{\tabcolsep}{4pt} 
\resizebox{1.0\linewidth}{!}{%
\begin{tabular}{p{0.05\linewidth} p{0.5\linewidth} ccc} 
\hline
\# & Methods & EMO-ACC \( (\uparrow) \) & MOS-DE \( (\uparrow) \) & MOS-SE \( (\uparrow) \) \\
\arrayrulecolor{black} \hline

\multicolumn{5}{l}{\textit{Effect of LLMs in Reference Footage Construction}} \\
\arrayrulecolor{gray!50} \hline
1 & w/o Scene Caption & 46.34 & 3.582 ± 0.032 & 3.612 ± 0.042 \\
2 & w/o Face Caption & 46.52 & 3.653 ± 0.049 & 3.684 ± 0.047 \\
3 & w/o Scene \& Face Caption & 46.02 & 3.520 ± 0.034 & 3.608 ± 0.053 \\
\arrayrulecolor{black} \hline
\multicolumn{5}{l}{\textit{Effect of Emotion-Similarity-based Retrieval-Augmentation}} \\
\arrayrulecolor{gray!50} \hline
4 & w/o Scene Retrieval & 46.27 & 3.591 ± 0.048 & 3.666 ± 0.045 \\
5 & w/o Face Retrieval & 46.64 & 3.657 ± 0.047 & 3.690 ± 0.051 \\
6 & w/o Text Retrieval & 45.99 & 3.540 ± 0.054 & 3.614 ± 0.047 \\
7 & w/o All Retrieval & 45.23 & 3.511 ± 0.058 & 3.527 ± 0.054 \\
\arrayrulecolor{black} \hline
\multicolumn{5}{l}{\textit{Effect of Progressive Graph-based Speech Generation}} \\
\arrayrulecolor{gray!50} \hline
8 & w/o Indirect Information & 45.95 & 3.542 ± 0.060 & 3.581 ± 0.053 \\
9 & w/o Direct Audio  & 45.30 & 3.492 ± 0.061 & 3.571 ± 0.051 \\
10 & w/o Graph-based Modeling & 45.92 & 3.518 ± 0.058 & 3.549 ± 0.055 \\
11 & w/o Construct \& Encode & 46.85 & 3.705 ± 0.049 & 3.749 ± 0.046 \\

12 & w/o Hierarchical Aggregation   & 46.71 & 3.661 ± 0.045 & 3.710 ± 0.050 \\
\arrayrulecolor{black} \hline
\end{tabular}%
}
\caption{Objective and subjective (with 95\% confidence interval) evaluation results of ablation studies.}
\label{table2}

\end{table}

\subsection{Comparative and Ablation Models}
To demonstrate the effectiveness of Authentic-Dubber in emotional expression, we compare it with five state-of-the-art dubbing models, namely FastSpeech \cite{fastspeech2}, V2C-Net \cite{V2CNET}, HPMDubbing \cite{HPMDubbing}, StyleDubber \cite{cong2024styledubber}, and Speaker2Dubber \cite{speaker2dubber}. In addition, to verify the contribution of each component in Authentic-Dubber, we conduct a comprehensive ablation study. Details on the baseline models and ablation settings can be found in Appendix A.

\subsection{Evaluation Metrics}

\textbf{Objective metrics.} (1) Emotion Accuracy (EMO-ACC): A pre-trained speech emotion recognition model \cite{TIM_NER_SER} is used to predict the emotion category of the synthesized speech and compute the percentage of matches with the ground-truth category. 
(2) Word Error Rate (WER): measures pronunciation accuracy by the Whisper-large-v3 automatic speech recognition model \cite{whisper}.
(3) Speaker Encoder Cosine similarity (SECS): Measures the speaker similarity between the synthesized speech and the timbre prompt following \cite{cong2024styledubber}. 

(4) Mel Cepstral Distortion Dynamic Time Warping weighted by Speech Length (MCD-DTW-SL) \cite{V2CNET}: Evaluates both length and the quality of alignment between the generated speech and the ground-truth speech.

\textbf{Subjective metrics.} We conducted a Mean Opinion Score (MOS) test with 20 trained raters who evaluated 12 generated
dubbed videos and speech samples, scoring the following metrics on a scale of 1 to 5 \cite{Multimodal-jia, hu2025unitalker, hu-etal-2025-chain, liu2024generative, jia2025intra}: (1) MOS-Dubbing Emotion (MOS-DE): Evaluates the emotional similarity between the generated dubbing and the ground-truth video.
(2) MOS-Speech Emotion (MOS-SE): Evaluate the emotional similarity between the synthesized speech and the ground-truth speech.

\section{Results and Discussion}

\subsection{Comparison with SOTA Dubbing Methods}

To verify the effectiveness of our method in enhancing emotional expressiveness, we compare the proposed model with several representative dubbing approaches. All speech samples generated by the baseline models are produced using their official implementations. As shown in Table \ref{table1}, our method achieves the best performance across all emotion-related evaluation metrics. Regarding objective measures, our approach obtains the highest EMO-ACC score (47.21\%), which indicates that our method is capable of generating speech with emotion more closely aligned with the ground truth. For subjective evaluations, our method also achieves the highest scores in both MOS-DE (3.792) and MOS-SE (3.889), indicating its superior ability to enhance the emotional expressiveness of both dubbed videos and generated speech. Overall, the results presented in Table \ref{table1} demonstrate the effectiveness of our method in enhancing emotional expression for dubbing by simulating the crucial interaction process between the director and the actors.

\begin{figure*}[tp]
    \centering
    \includegraphics[width=\linewidth]{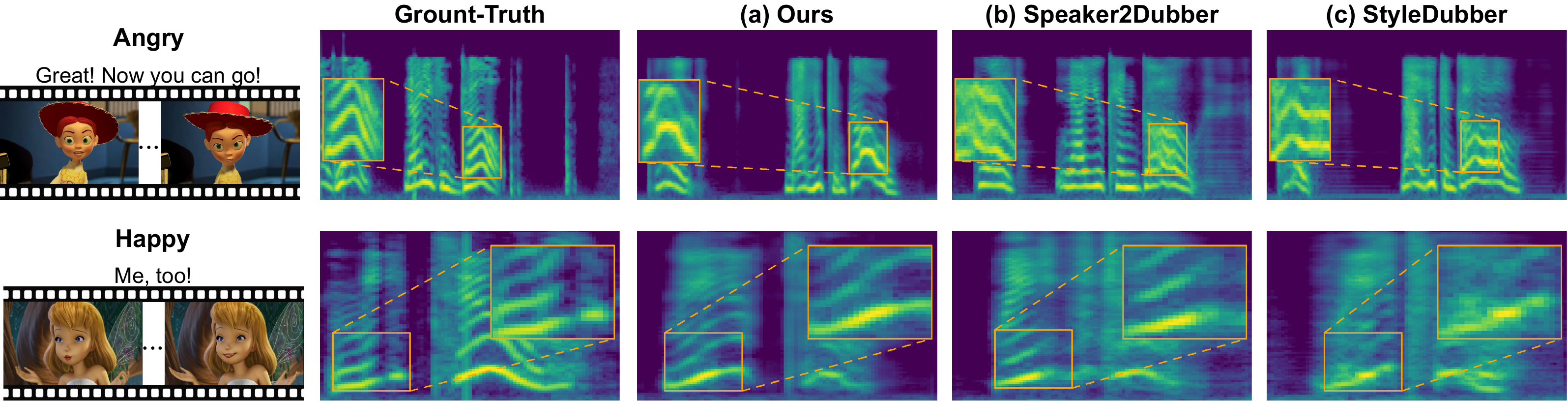}
    \caption{The visualization of the mel-spectrograms of ground truth (GT) and synthesized speech obtained by different dubbing baselines, and orange bounding boxes are used to highlight the details in speech.}
    \label{Visualization}
\end{figure*}

\begin{figure}[t]
    \centering
    \includegraphics[width=\linewidth]{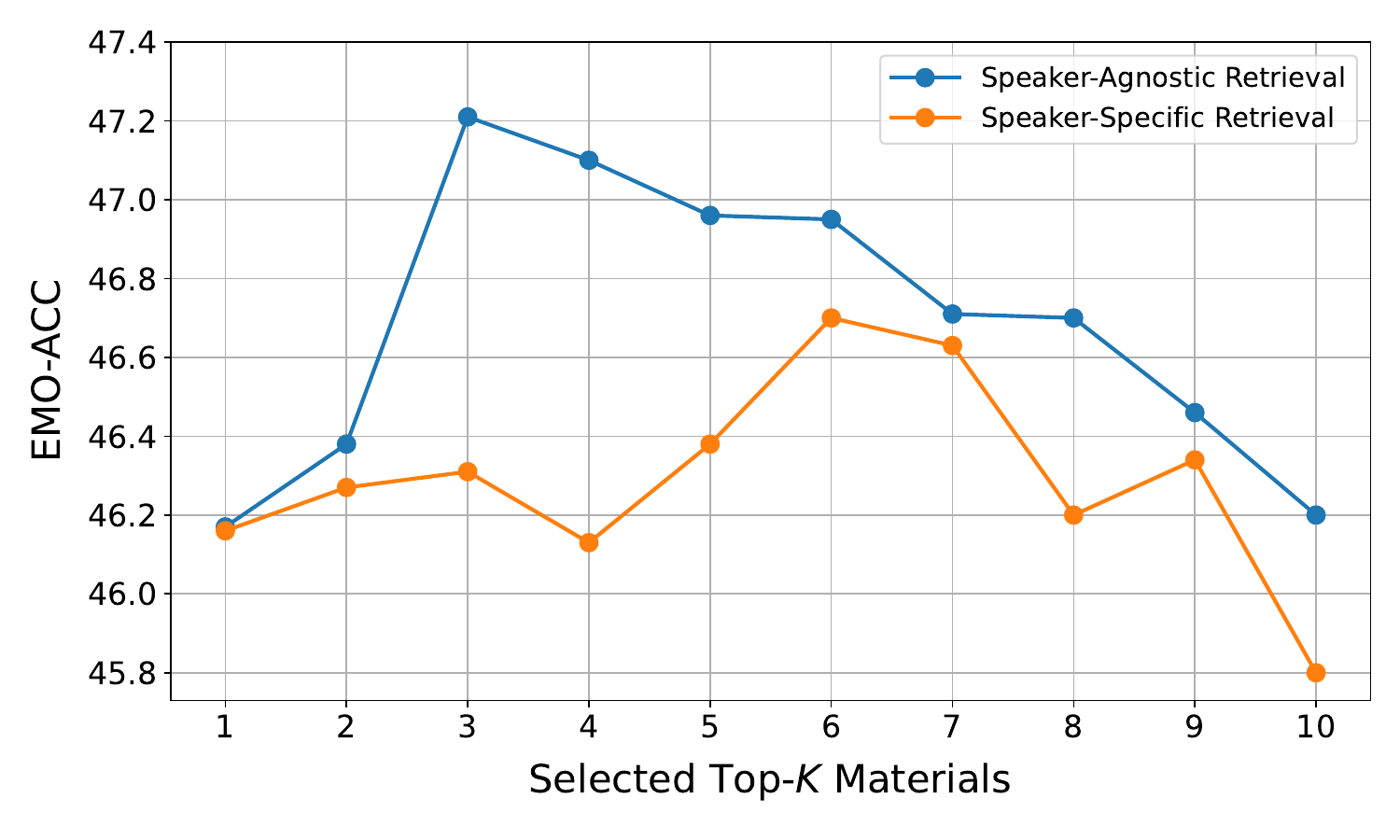}
    \caption{EMO-ACC of speech generated by Authentic-Dubber under Speaker-Agnostic and Speaker-Specific retrieval settings with varying Top-$K$ values.}
    \label{fig_Top_k}
\end{figure}

\begin{figure}[t]
    \centering
    \includegraphics[width=\linewidth]{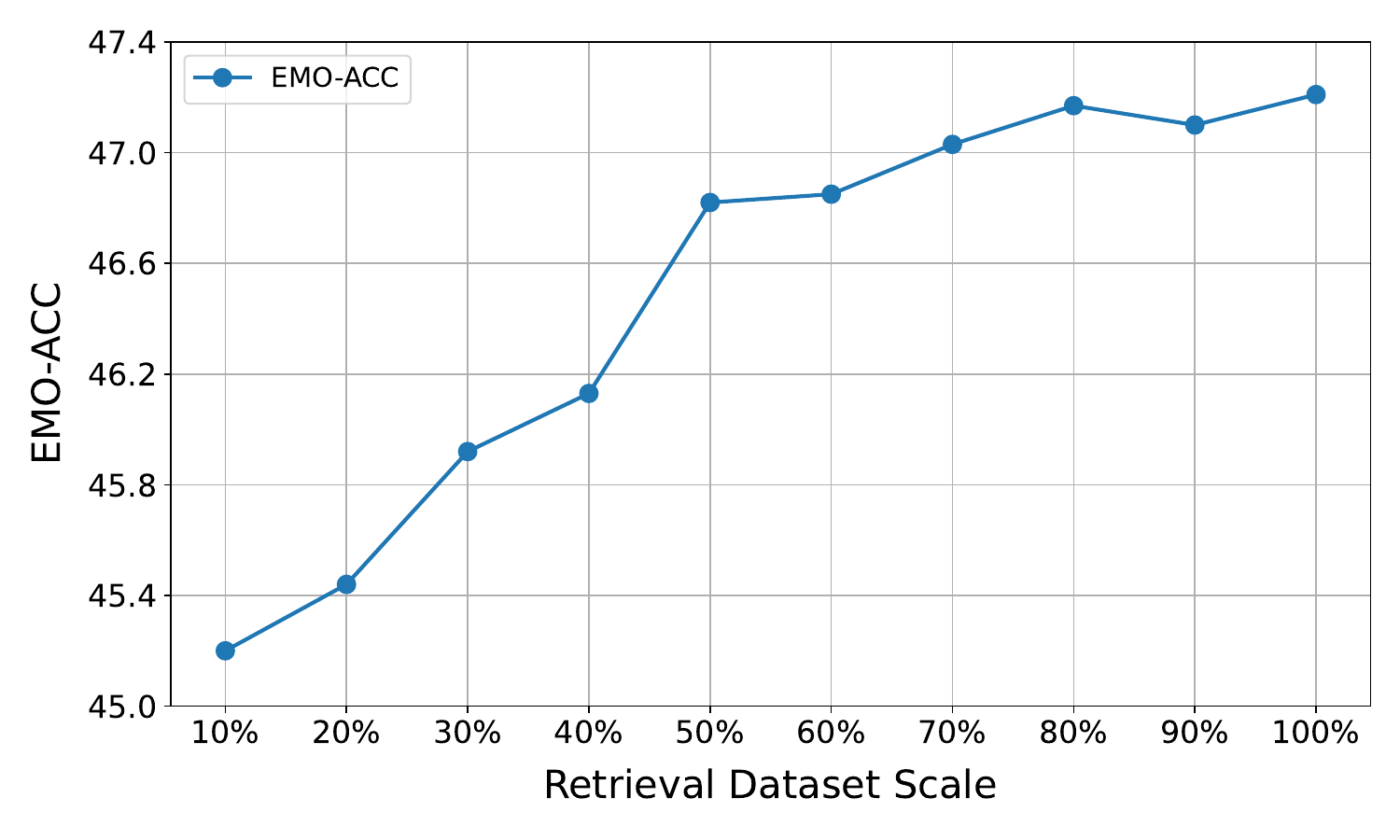}
    \caption{The figure illustrates the emotion accuracy (EMO-ACC) of speech generated by our proposed Authentic-Dubber under different retrieval dataset scales.}
    \label{dataset_scale}
\end{figure}

\begin{figure}[t]
    \centering
    \includegraphics[width=\linewidth]{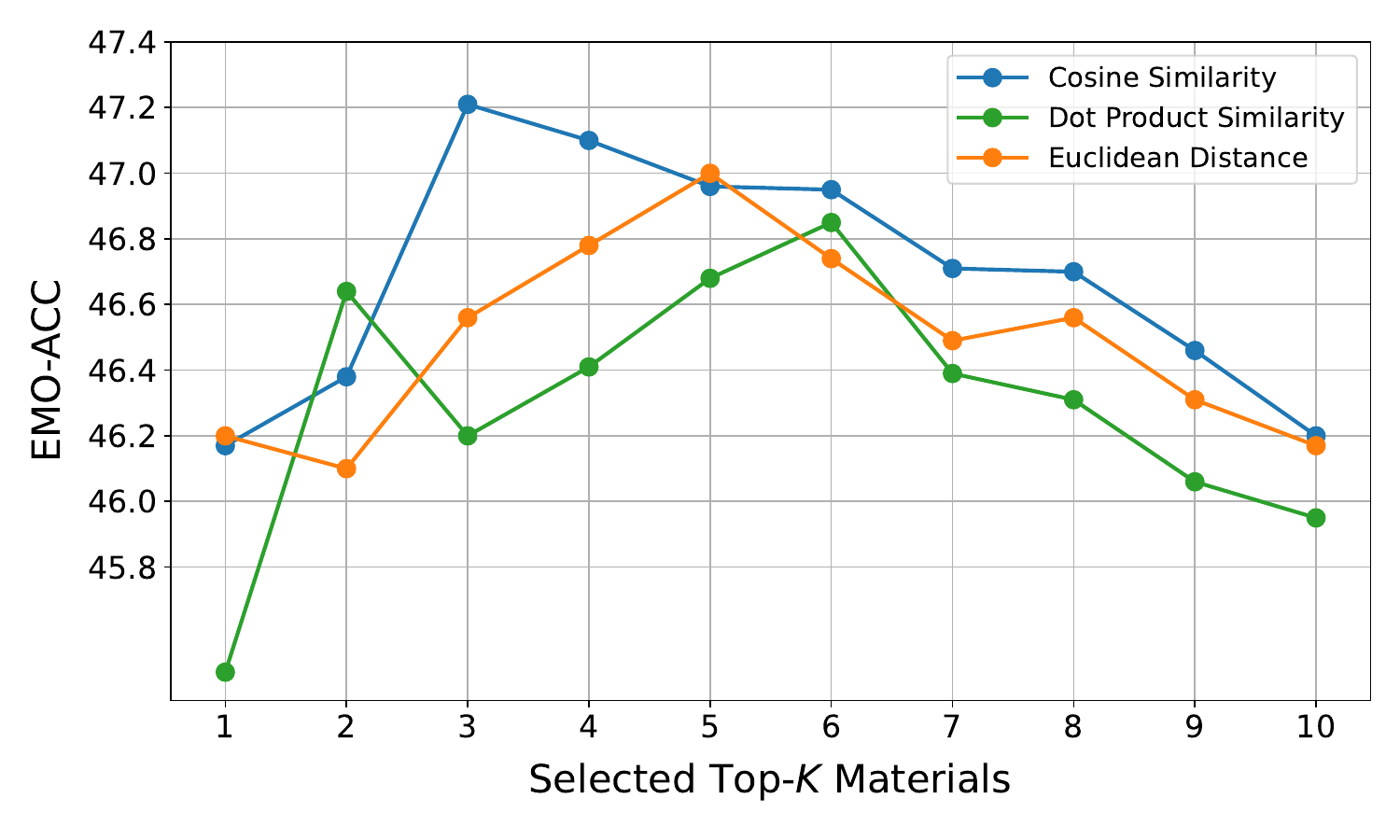}
    \caption{The figure shows the emotion accuracy (EMO-ACC) of speech generated by our proposed Authentic-Dubber using different vector similarity metrics during retrieval.}
    \label{different_compute}
\end{figure}

\subsection{Ablation Studies}
To validate the contribution of each component in Authentic-Dubber, we performed comprehensive ablation studies (Table \ref{table2}). Specifically, to validate the \textit{Effect of LLMs in Reference Footage Construction}, We conducted ablations by replacing the Scene Caption with embeddings extracted from the scene emotion model I3D \cite{I3D_model}, replacing the Face Caption with embeddings from the facial emotion model EmoFan \cite{emofan}, and replacing both captions simultaneously, as shown in Table \ref{table2}, lines 1–3. The results show a consistent performance drop across all evaluation metrics, demonstrating that the LLM’s deep comprehension of emotional representations across multimodal signals effectively contributes to the model’s performance by enhancing its emotional modeling capability. To validate the \textit{effectiveness of Emotion-Similarity-based Retrieval-Augmentation}, we performed ablations by individually removing scene retrieval, face retrieval, text retrieval, and all retrieval, as shown in Table \ref{table2}, lines 4–7. The results show that all evaluation metrics dropped, with the most significant decline observed when all retrievals were removed. This indicates that each retrieval modality effectively contributes to emotional information enhancement, and their combined use plays a crucial role in providing rich emotional cues, thereby improving the overall dubbing performance. In addition, to validate the \textit{Effect of Emotion Knowledge-based Speech Generation}, we ablate key components including Indirect Information, Direct Audio, Graph-based Modeling, Construct \& Encode, and Hierarchical Aggregation, as shown in Table \ref{table2}, lines 8–12. The results show that all evaluation metrics decline, further demonstrating that the basic emotion, indirect multimodal emotional information, direct emotional audio, as well as the construct-and-encode paradigm and hierarchical aggregation strategy, directly affect the emotional expressiveness and speech quality of the generated dubbing. Moreover, this paradigm of progressively integrating diverse emotional knowledge through graph structures proves to be crucial for producing emotionally expressive dubbing speech.

\subsection{Speaker-Agnostic vs. Speaker-Specific Retrieval}

This experiment evaluates how the retrieval setting (Speaker-Agnostic vs. Speaker-Specific) affects the emotional accuracy (EMO-ACC) of generated speech. As shown in Fig. \ref{fig_Top_k}, Speaker-Agnostic retrieval achieves the highest EMO-ACC (47.21\%) at $K$ = 3 and consistently outperforms the Speaker-Specific setting. This supports our strategy of using speaker-agnostic retrieval to enrich the selected reference footage in animation dubbing scenarios with few speaker priors, thereby enhancing emotional expressiveness. However, increasing $K$ beyond a certain point degrades performance under both settings, suggesting that excessive retrieval introduces redundant information regardless of the retrieval strategy.

\subsection{Qualitative Analysis of Mel-Spectrogram}

This experiment compares our model with several baseline methods by visualizing the mel-spectrograms of generated speech in two emotional categories: angry and happy. As shown in Fig. \ref{Visualization}, the zoomed-in regions highlighted with orange bounding boxes indicate that our model more accurately captures high fluctuations in angry speech and exhibits more natural prosodic variation in happy speech. These results suggest that our model conveys emotional expressiveness more effectively than the baselines.

\subsection{Analysis of Retrieval Footage Scales}

To assess the effect of retrieval dataset scale, that are reference footage scale, on emotional expressiveness, we evaluated our model’s EMO-ACC across scales from 10\% to 100\%. As shown in Fig.~\ref{dataset_scale}, EMO-ACC steadily increases with larger datasets, plateauing between 80\% and 100\%, with a peak at 47.21\%. This demonstrates that expanding the emotion footage set enhances emotional expressiveness.

\subsection{Analysis of Similarity Metrics}

To investigate the impact of vector similarity metrics in retrieval on model performance in terms of EMO-ACC, we conducted comparative experiments evaluating cosine similarity, dot product, and Euclidean distance under different Top-$K$ settings. As shown in Fig. \ref{different_compute}, cosine similarity consistently yields the best overall performance. Among the metrics, the dot product exhibits greater fluctuations, while Euclidean distance is relatively stable but shows a slightly lower performance ceiling. These findings indicate that the model is sensitive to the choice of similarity function and that retrieving a small number of high-quality emotional footage is more effective for enhancing the emotional expressiveness of synthesized speech.

\section{Conclusion}

In this paper, we propose \textit{Authentic-Dubber}, a framework that simulates real-world dubbing workflows to enhance the emotional expressiveness of movie dubbing. A multimodal Reference Footage Library is constructed to simulate the learning materials typically provided by directors. An Emotion-Similarity-based Retrieval-Augmentation strategy is designed to retrieve the most relevant multimodal information aligned with the target silent video. A Progressive Graph-based Speech Generation module is developed to incrementally incorporate the retrieved emotional knowledge. Experimental results demonstrate the effectiveness of our model on the V2C benchmark. In future work, we plan to explore more expressive modeling of director–actor interactions in real dubbing workflows, including controllable aspects such as timbre, speaking rate, and prosody.

\section{Acknowledgments}

This research was funded by the Young Scientists Fund (No. 62206136) and the General Program (No. 62476146) of the National Natural Science Foundation of China, the Young Elite Scientists Sponsorship Program by CAST (2024QNRC001), the Outstanding Youth Project of Inner Mongolia Natural Science Foundation (2025JQ011), the Key R\&D and Achievement Transformation Program of Inner Mongolia Autonomous Region (2025YFHH0014), and the Central Government Fund for Promoting Local Scientific and Technological Development (2025ZY0143).

\bibliography{aaai2026}

\end{document}